\documentclass[sigconf, nonacm]{acmart}
\AtBeginDocument{%
  }

\usepackage{algorithm}
\usepackage{booktabs}

\usepackage{algpseudocode}

\usepackage{stfloats}
\begin{document}

\title{AdaGradSelect: An adaptive gradient-guided layer selection method for efficient fine-tuning of SLMs }


\author{Anshul Kumar}
\email{anshulchrs@gmail.com}
\orcid{ 0009-0001-4230-8877}
\affiliation{%
  \institution{IIT Bhilai}
  \country{India}
  }

\author{Gagan Raj Gupta}
\email{gagan@iitbhilai.ac.in}
\orcid{0000-0002-8568-2949}
\affiliation{%
  \institution{IIT Bhilai}
  \country{India}
}

\author{Manisha Chawla}
\email{manishachawla@iitbhilai.ac.in}
\affiliation{%
  \institution{IIT Bhilai}
  \country{India}
  }


\begin{abstract} 
While Large Language Models (LLMs) excel at diverse NLP tasks, their adaptation through full fine-tuning is computationally expensive and memory-intensive. Parameter-Efficient Fine-Tuning (PEFT) methods like Low-Rank Adaptation (LoRA) mitigate this by introducing low-rank updates to frozen weights, but this constrains optimization to a low-rank subspace and can limit performance. Focusing on Small Language Models (SLMs), where efficiency gains offer significant practical benefits, we introduce \textbf{AdaGradSelect}, an adaptive, gradient-guided block selection strategy for efficient fine-tuning. 

Motivated by preliminary findings that selectively updating transformer blocks with the highest gradient norms approaches full fine-tuning performance, AdaGradSelect dynamically prioritizes which blocks to train. The method combines Dirichlet-based sampling, informed by historical update frequencies, with an $\epsilon$-greedy exploration strategy. This approach initially balances the exploitation of important blocks with the exploration of new candidates before transitioning to full exploitation in later epochs, optimizing the training process. 

Experimental results demonstrate that AdaGradSelect trains approximately 12\% faster and uses 35\% less GPU memory while achieving performance nearly identical to full fine-tuning. On the GSM8K dataset, our method consistently outperforms LoRA (rank 256) by an average of 3\% across Qwen2.5-0.5B, LLaMA3.2-1B, and Phi4-mini-3.8B models. It also shows comparable accuracy on the MATH dataset, establishing AdaGradSelect as a more effective and resource-efficient fine-tuning approach. 
\end{abstract}




\maketitle
\section{Introduction}

Large Language Models (LLMs) have demonstrated exceptional performance across a wide range of tasks, including documentation generation, complex code synthesis, question answering, and human-like conversation \cite{ouyang2022traininglanguagemodelsfollow}. Their success on diverse NLP benchmarks \cite{fourrier2025openllm} is largely attributed to strong generalization and capabilities such as in-context learning. However, LLMs often underperform on specialized or low-resource tasks, necessitating domain-specific fine-tuning to bridge the performance gap \cite{raffel2020exploring, roziere2023code}.

Despite its effectiveness, full fine-tuning of LLMs is computationally expensive and memory-intensive. The memory footprint is dominated not only by billions of model parameters but also by additional optimizer states and gradients, such as momentum and variance terms in Adam, which can exceed the storage requirements of the parameters themselves \cite{touvron2023llama, chowdhery2023palm}. These constraints make full fine-tuning prohibitively costly for most practical applications. A practical alternative is to focus on Small Language Models (SLMs) \cite{slm}. They are much smaller than LLMs, but can still be fine-tuned effectively. They are also easier to deploy in environments with limited resources. In this work, we go a step further. Our aim is to make fine-tuning even faster and more effective, so that SLMs provide the maximum possible benefit in practice.

\begin{figure}[H]
    \centering
    \includegraphics[width=\columnwidth]{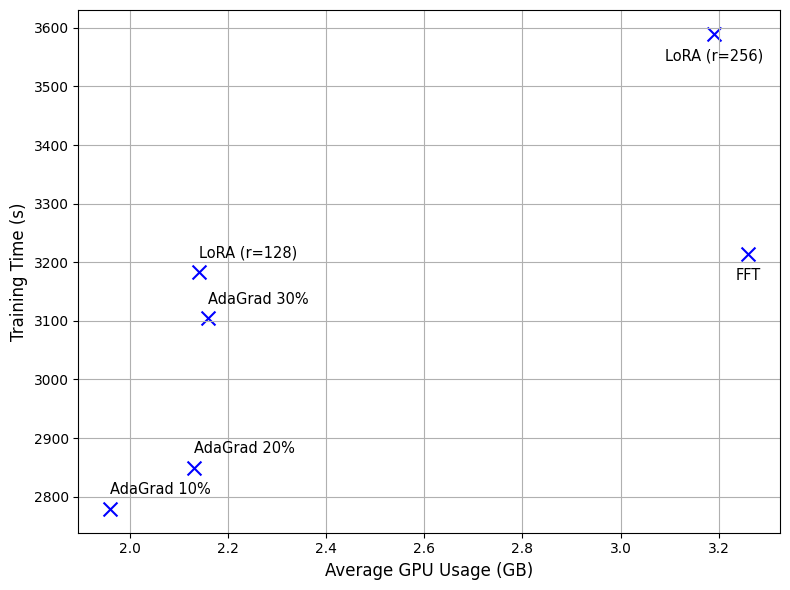}
    \caption{Comparison of training time vs Avg GPU usage for training Qwen2.5 0.5B using different methods}
    \label{fig:gpu_vs_time}
\end{figure}

\textbf{Parameter Efficient Fine Tuning(PEFT)} methods have emerged as a promising alternative, allowing adaptation of pre-trained language models to downstream tasks without updating all parameters \cite{ding2022delta}. Among these, \textbf{Low-Rank Adaptation (LoRA)} \cite{hu2021loralowrankadaptationlarge} is widely adopted. LoRA reparameterizes a weight matrix $W \in \mathbb{R}^{m \times n}$ as:
\[
W = W_0 + BA,
\]
where $W_0$ is a frozen pre-trained matrix, and $B \in \mathbb{R}^{m \times r}$, $A \in \mathbb{R}^{r \times n}$ are learnable low-rank adapters with $r \ll \min(m,n)$. This significantly reduces trainable parameters and optimizer states, lowering memory usage. While LoRA achieves efficiency gains, its optimization is constrained to a low-rank subspace, which can limit representational power and prevent it from matching full fine-tuning performance in many cases \cite{ding2022delta, xia2024chain}.And the case of SLMs with small hidden dimensions, adapter-based methods offer limited benefits and can even result in longer training times compared to full fine-tuning, as shown in Figure~\ref{fig:gpu_vs_time}. This occurs because the additional overhead from the adapter forward pass, combined with the computation of the model’s original weights, becomes significant in smaller models. While such effects are largely mitigated in larger models, they remain noticeable in smaller ones.


To address these limitations, we introduce \textbf{AdaGradSelect} — an adaptive gradient-guided block selection strategy for efficient LLM fine-tuning (custom AdamW). Instead of updating all transformer blocks, AdaGradSelect selectively updates a subset of high-impact blocks determined by cumulative gradient norms. We enhance this approach with:
\begin{itemize}
    \item \textbf{Dirichlet-based sampling}, which biases block selection toward historically significant blocks using update frequency counts.
    \item An \textbf{$\epsilon$-greedy exploration strategy} during the first epoch, enabling controlled exploration of alternative blocks. The exploration probability decays exponentially within the first epoch, after which the process transitions to pure exploitation.
\end{itemize}
This formulation draws parallels with the Multi-Armed Bandit problem, where blocks act as arms, and the goal is to maximize adaptation efficiency under resource constraints.

\textbf{Our contributions are as follows:}
\begin{enumerate}
    \item We empirically analyze the effect of fine-tuning only a subset of transformer blocks based on gradient norms, showing that updating as few as 10\% of blocks can be comparable to full fine-tuning performance.
    \item We propose \textbf{AdaGradSelect}, a novel block selection algorithm that combines Dirichlet-based exploitation with $\epsilon$-greedy exploration to reduce computational overhead and memory usage without compromising accuracy.
    \item Through extensive experiments, we show that AdaGradSelect consistently outperforms LoRA with rank 256 by an \textbf{average of 3\% on the GSM8K dataset} across Qwen2.5-0.5B, LLaMA3.2-1B, and Phi4-mini-3.8B models, while achieving comparable accuracy on the MetaMathQA-40K dataset, establishing it as both \textbf{more effective and more resource-efficient} than existing fine-tuning methods.
    \item Furthermore, our method demonstrates clear efficiency gains, achieving approximately 12\% faster training and 35\% lower GPU memory usage, while maintaining nearly the same accuracy as full fine-tuning as shown in Figure \ref{fig:gpu_vs_time}.

\end{enumerate}


\section{Related Work}

LLMs such as GPT-3\cite{brown2020language}, PaLM\cite{chowdhery2022palm}, LLaMA\cite{touvron2023llama}, and T5\cite{raffel2020exploring} have achieved state-of-the-art performance across a broad spectrum of natural language processing tasks. Traditionally, adapting these models to downstream tasks has relied on full fine-tuning, wherein all model parameters are updated. Although effective, full fine-tuning is computationally intensive and memory-expensive, making it infeasible for most resource-constrained environments.

To overcome these limitations, recent research has shifted attention toward SLMs, compact but capable variants of LLMs. SLMs offer significant practical benefits, such as they are faster to train, cheaper to deploy, and more suitable for edge devices, low-resource environments, and specialized domains \cite{drllma,flame, fingpt, lawyerllamatechnicalreport}. Despite their smaller size, they remain competitive across a range of NLP tasks \cite{qwen,phi}. This makes them an attractive alternative to LLMs, especially when efficient fine-tuning strategies can further unlock their potential. Our work is motivated by this direction, focusing on improving the speed and effectiveness of SLM fine-tuning to maximize their impact in practical settings.

Beyond this, researchers have also explored Parameter-Efficient Fine-Tuning (PEFT) methods to reduce the cost of adapting large pre-trained models. Representative approaches include Low-Rank Adaptation (LoRA) \cite{hu2021lora}, which inserts trainable low-rank matrices into attention modules; Adapter Tuning \cite{houlsby2019parameter}, which introduces lightweight modules between transformer blocks; Prefix-Tuning \cite{li2021prefix}, which prepends continuous task-specific vectors to the input; and BitFit \cite{zaken2022bitfit}, which updates only bias terms. These methods significantly reduce the number of trainable parameters, but they often constrain optimization to a limited subspace, which can result in suboptimal performance compared to full fine-tuning.

Recent methods attempt to bridge this gap by exploring gradient- or structure-based selective updates. For example, GaLore\cite{zhao2024galore} reduces memory usage during training by projecting gradients into a low-rank subspace. While effective, it still requires access to gradients and may not reach full fine-tuning performance. Similarly, LISA\cite{pan2024lisa} introduces a layerwise importance sampling strategy, freezing most intermediate layers and selectively updating only a few high-impact layers. LISA matches or even outperforms LoRA and full fine-tuning while using comparable memory, showcasing the promise of selective updates in LLM training.

Other approaches such as Diff-Pruning\cite{guo2021parameter} and LayerDrop\cite{fan2019reducing} encourage sparse or partial updates to transformer parameters. AutoFreeze\cite{liu2021autofreeze}, SmartFRZ\cite{li2023smartfrz}, and FreezeOut\cite{brock2017freezeout} attempt to identify and freeze redundant layers, although these methods often involve complex heuristics or architectural changes, limiting their practicality in large-scale LLM settings.

In contrast, our method is inspired by a simple but powerful observation that only a small subset of transformer blocks significantly contributes to adaptation during fine-tuning. Our preliminary experiments using Gradient-Guided Block Selection confirm this behavior and motivate our proposed approach AdaGradSelect. Unlike prior methods, AdaGradSelect tracks update frequencies and employs Dirichlet-based sampling to prioritize impactful blocks. An $\epsilon$-greedy exploration ensures diverse early updates, while convergence naturally focuses on the most relevant blocks. Crucially, AdaGradSelect avoids gradient access and architectural modifications, enabling scalable and efficient fine-tuning in black-box or low-resource LLM environments.

\begin{figure*}[t!]
    \centering
    \includegraphics[width=\textwidth, height=0.42\textheight, keepaspectratio]{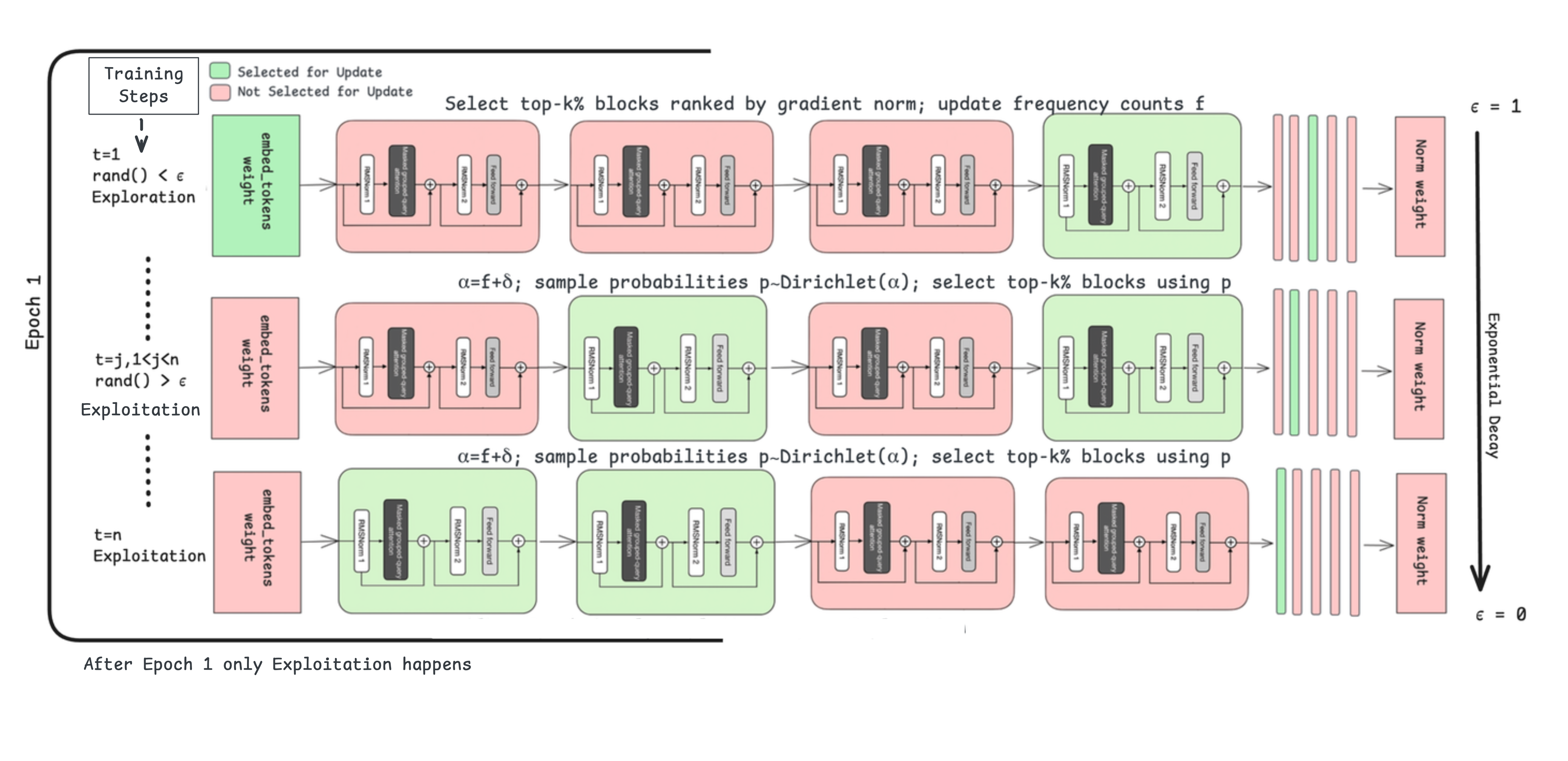}
    \caption{Illustration of AdaGradSelect’s selective block update strategy. During fine-tuning, only a subset of transformer blocks (green) are updated, while others remain frozen (red). In the first epoch, an $\epsilon$-greedy strategy enables exploration-exploitation as shown, with $\epsilon$ decaying exponentially. At first step there will always be exploration and at Nth step there will always be Exploitatio After Epoch~1, the method transitions fully to exploitation.}

    \label{fig:arch}
\end{figure*}

\section{Methodology}

We begin by conducting a preliminary experiment to evaluate the effect of selectively updating only the top-k\% transformer blocks with the highest gradient norms. The promising results, which approach full fine-tuning performance while reducing computational cost, motivate the design of our proposed method.

Building on this idea, and to reduce the overhead from calculating and ranking blocks by gradient norm, we introduce AdaGradSelect, an optimized strategy formulated as a multi-armed bandit problem. Specifically, we adopt an $\epsilon$-greedy approach: during the exploration phase, block selection is guided by cumulative gradient norms, while in the exploitation phase, frequency counts of past selections drive block updates (Figure~\ref{fig:arch}).

\subsection{Motivation}
Full fine-tuning of all transformer blocks in language model is a computationally intensive and resource-demanding process. To investigate whether a selective fine-tuning approach, focusing on a subset of transformer blocks, can achieve comparable performance to full fine-tuning while significantly reducing training costs, we conducted a series of empirical experiments. Our methodology is designed to systematically evaluate the trade-offs between computational efficiency and model performance under various selective update configurations.

We define a "block" as a collection of layers within the transformer architecture with embed weights and final norm weight also as blocks other than the transformer blocks which are individual layers, the transformer blocks encompass all layers like the multi-head attention and feed-forward network components, basically everything inside the current modern transformer blocks. To prioritize which blocks are most critical for adaptation and thus should be updated, we employed a gradient-guided selection mechanism. This involves computing the L2 norm of the gradients for each parameter, aggregating these norms block-wise, and subsequently selecting the top k \% of blocks exhibiting the highest cumulative gradient norms and only update them as shown in Figure \ref{fig:arch}., We choose percentage cause it adapts to size of model and gives better understanding, rather than fixing number of blocks.. The detailed procedure for this selective block update is outlined in Algorithm ~\ref{alg:selective-block-update}:

\begin{algorithm}
\caption{Gradient Guided Layer Selection (Selective Block Update)}
\label{alg:selective-block-update}
\begin{algorithmic}[1]  
\State Initialize $\texttt{block\_norm}[b] \gets 0$ for each block $b$
\For{each $\texttt{weight} \in \texttt{model.parameters()}$}
    \State $\texttt{grad} \gets \texttt{weight.grad}$
    \State $\texttt{block} \gets$ block index of \texttt{weight}
    \State $\texttt{block\_norm[block]} \mathrel{+}= \|\texttt{grad}\|$
\EndFor
\State $\texttt{sorted\_blocks} \gets$ sort blocks by descending $\texttt{block\_norm}$
\State $\texttt{selected\_blocks} \gets$ top-$k$\% of $\texttt{sorted\_blocks}$
\For{each $\texttt{weight} \in \texttt{model.parameters()}$}
    \If{\texttt{weight} is in a block $\in \texttt{selected\_blocks}$}
        \State $\texttt{weight.data} \mathrel{+}= \texttt{update}$
    \EndIf
\EndFor
\end{algorithmic}
\end{algorithm}

For our experiments, we utilized Qwen2.5-0.5B as the base pre-trained language model. This model comprises 25 transformer blocks. Fine-tuning was performed on the MetaMath40K dataset, a specialized dataset designed for mathematical reasoning, consisting of 40,000 high-quality problems and solutions. Model evaluation was conducted on the GSM8K test set, a widely recognized benchmark for grade school math problems, which provides a robust measure of arithmetic and logical reasoning capabilities. The primary objective of these experiments was to systematically compare training time, final model accuracy on GSM8K, and overall computational efficiency across various percentages of updated blocks, ranging from selective updates to full fine-tuning.

\begin{figure}[H]
    \centering
    \includegraphics[width=\columnwidth]{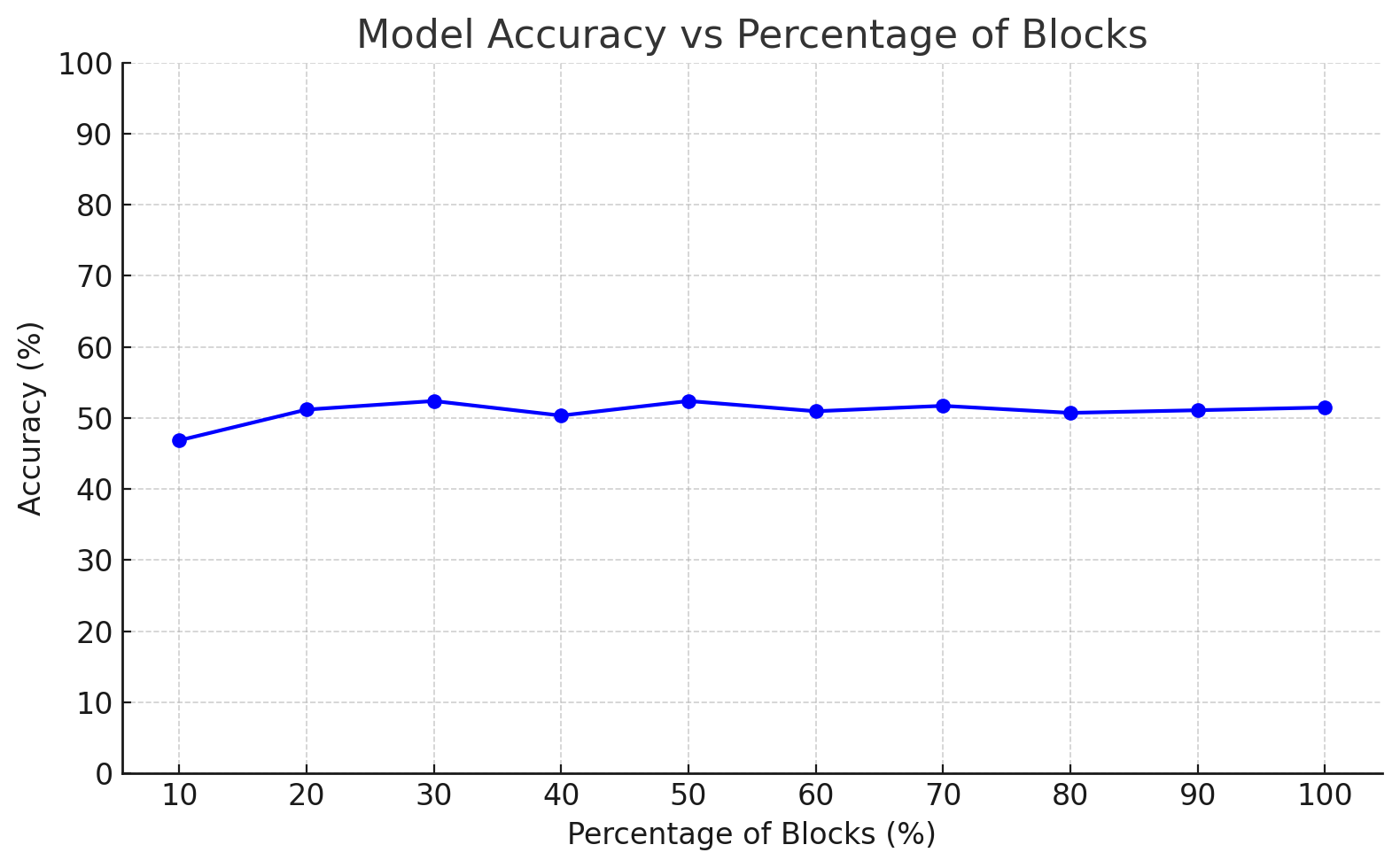}
    \caption{Comparison of Accuracy vs Percentage of Qwen2.5 0.5B Transformer Blocks Selected}
    \label{fig:acc_vs_per}
\end{figure}

Figure ~\ref{fig:acc_vs_per} visually summarizes the impact of updating different percentages of transformer blocks on GSM8K accuracy. The baseline for comparison corresponds to full fine-tuning (FFT) of all 25 blocks, representing the standard approach.

Our initial experimental findings reveal a significant insight: fine-tuning as few as 10\% of the transformer blocks (specifically, 2 out of the 25 blocks in Qwen2.5-0.5B) achieves performance comparable to that of full fine-tuning. This selective approach concurrently reduces the total training time by approximately 15\%. This finding underscores the potential for substantial efficiency gains without a significant compromise in model accuracy, suggesting that only the most impactful blocks require adaptation during fine-tuning.

This line of experimentation draws conceptual inspiration from the Lottery Ticket Hypothesis in neural network pruning, which posits the existence of certain sparse subnetworks within a larger model that, when trained in isolation, can achieve performance comparable to the full, dense model. Analogously, our results suggest that a small, strategically selected subset of high-gradient transformer blocks can effectively serve as an "effective ticket" for fine-tuning, thereby reducing computational overhead while preserving the model's performance. Language model are architecturally composed of multiple stacked transformer blocks, in addition to embedding layers and final normalization layers. Each transformer block contributes to building contextual representations of the input embeddings through self-attention mechanisms. While these blocks typically adapt to domain-specific data during fine-tuning or even with parameter-efficient methods like LoRA, our experiments demonstrate that not all blocks necessitate such adaptation. Given that the output of one block serves as the input to the subsequent block, if a limited subset of blocks effectively learns the essential contextual information from the fine-tuning dataset, the overall model can achieve robust adaptation to the new domain.

To further validate this premise, we intentionally selected a smaller parameter model (Qwen2.5-0.5B), which inherently exhibits poor performance on domain-specific tasks without fine-tuning. Interestingly, our observations consistently indicated that the initial layers of the model received the largest proportion of updates when employing selective gradient-guided fine-tuning. Conversely, deeper transformer blocks exhibited significantly fewer updates. This pattern suggests a hierarchical importance among blocks, where early layers contribute more prominently to the process of domain adaptation.

We conducted a detailed analysis of the update frequency distribution of transformer blocks across configurations involving less than 50\% block selection, as these configurations yield the most significant efficiency gains. Our analysis consistently revealed a persistent pattern: a few specific blocks, predominantly the initial layers, are updated far more frequently than others. Moreover, the dominance of these early blocks is so pronounced that even when additional blocks are introduced later in the training process, the overall update distribution remains largely unchanged. This observed stability further reinforces our premise that a small, critical subset of blocks is sufficient to capture the essential domain-specific information required for effective and efficient fine-tuning.

\subsection{AdaGradSelect}

We propose \textbf{AdaGradSelect}, an adaptive block selection strategy designed to improve the efficiency of fine-tuning Language Models. The method prioritizes high-impact transformer blocks using cumulative gradient norms and introduces an adaptive exploration-exploitation mechanism grounded in probabilistic sampling. This approach reduces training cost while preserving performance comparable to full fine-tuning.

\paragraph{Dirichlet-Based Sampling.}
To exploit historical information about block importance, we model selection probabilities using a Dirichlet distribution. Let $f_i$ denote the frequency of updates for block $i$ observed during training. The Dirichlet parameter vector $\boldsymbol{\alpha}$ is defined as:
\[
\alpha_i = f_i + \delta,
\]
where $\delta > 0$ is a smoothing constant ensuring non-zero probabilities. At each selection step, we sample a probability vector $\mathbf{p} \sim \text{Dirichlet}(\boldsymbol{\alpha})$ and select the top-$k$\% blocks by drawing without replacement according to $\mathbf{p}$. This stochastic process favors historically impactful blocks while maintaining diversity in selection.

\begin{algorithm}[H]
\caption{AdaGradSelect: Adaptive Block Selection with Dirichlet Sampling and $\epsilon$-Greedy}
\label{alg:adagardselect}
\begin{algorithmic}[1]  
\State Initialize frequency counts $\mathbf{f} \gets 0$
\For{each training step}
    \If{epoch == 1}
        \If{rand() < $\epsilon$} \Comment{Exploration}
            \State selected\_blocks $\gets$ top-$k$\% by gradient norm
        \Else \Comment{Exploitation}
            \State $\boldsymbol{\alpha} \gets \mathbf{f} + \delta$
            \State $\mathbf{p} \sim \text{Dirichlet}(\boldsymbol{\alpha})$
            \State selected\_blocks $\gets$ sample $k$\% using $\mathbf{p}$
        \EndIf
        \State $\epsilon \gets \epsilon_0 e^{-\lambda t}$ \Comment{Exponential decay}
    \Else \Comment{Epoch $\geq$ 2: Pure exploitation}
        \State $\boldsymbol{\alpha} \gets \mathbf{f} + \delta$
        \State $\mathbf{p} \sim \text{Dirichlet}(\boldsymbol{\alpha})$
        \State selected\_blocks $\gets$ sample $k$\% using $\mathbf{p}$
    \EndIf
    \State Update frequency counts for selected\_blocks
    \State Update model parameters in selected\_blocks
\EndFor
\end{algorithmic}
\end{algorithm}

\paragraph{Exploration-Exploitation Dynamics.}
In AdaGradSelect, exploration and exploitation are tightly coupled through an adaptive feedback loop. During the first epoch, at each step:
\begin{itemize}
    \item With probability $(1 - \epsilon)$, the algorithm performs \textbf{exploitation} by sampling blocks from the Dirichlet distribution based on current frequency counts $\mathbf{f}$.
    \item With probability $\epsilon$, the algorithm performs \textbf{exploration} by selecting the current top-$k$\% blocks ranked by gradient norms.
\end{itemize}
After every selection, frequency counts $\mathbf{f}$ are updated, ensuring that exploration influences future exploitation by shaping the Dirichlet distribution. The exploration probability $\epsilon$ decays exponentially as:
\[
\epsilon_t = \epsilon_0 e^{-\lambda t},
\]
where $t$ is the training step, $\epsilon_0$ is the initial exploration rate, and $\lambda$ controls the decay. From epoch 2 onward, exploration is disabled ($\epsilon = 0$), and block selection relies purely on Dirichlet-based exploitation using the accumulated frequencies .

\paragraph{Connection to Multi-Armed Bandit.}
Block selection can be framed as a Multi-Armed Bandit (MAB) problem, where each transformer block is an arm and the objective is to maximize cumulative reward (e.g., accuracy gains) under resource constraints. AdaGradSelect aligns with MAB principles: Dirichlet-based sampling reflects prior knowledge through frequency counts, while $\epsilon$-greedy ensures sufficient exploration of untested blocks early in training.

\paragraph{Training Phases.}
\begin{itemize}
    \item \textbf{Epoch 1 (Exploration-Exploitation):} Apply Dirichlet sampling combined with $\epsilon$-greedy exploration. $\epsilon$ decays exponentially throughout this phase.
    \item \textbf{Epoch $\geq$ 2 (Exploitation):} Selection is driven purely by Dirichlet sampling based on accumulated frequencies.
\end{itemize}



This adaptive design ensures that early exploration informs later exploitation, progressively refining block selection as training evolves. Empirical evaluations demonstrate that AdaGradSelect achieves substantial reductions in training time and GPU memory usage while maintaining accuracy on par with full fine-tuning.



\subsection{GPU Optimization}
Since AdaGradSelect updates only the parameters of selected blocks, this creates an opportunity for GPU memory optimization. We leverage this by implementing a dynamic optimizer state management strategy. All optimizer states, such as the momentum and variance accumulators in AdamW, are initially stored in CPU RAM. At each training step, optimizer states for newly selected blocks are asynchronously prefetched from CPU to GPU, while states for blocks no longer selected are evicted back to CPU. States for blocks that remain selected across consecutive steps stay resident on the GPU, avoiding redundant transfers. This asynchronous prefetch-and-evict mechanism ensures that, at any point, only the optimizer states for the actively updated portion of the model occupy VRAM. As a result, GPU memory overhead from the optimizer is reduced to a fraction of its full-model requirement, enabling the training of larger models or the use of larger batch sizes.

\paragraph{Optimizer State Memory Calculation}  
The GPU memory used by the AdamW optimizer states can be estimated as:
\[
\text{Mem}_{\text{Optimizer}} = 2 \times (\text{\# Parameters on GPU}) \times (\text{Bytes per Parameter})
\]
where the factor of $2$ accounts for the momentum and variance accumulators.

Let:
\begin{itemize}
    \item $P_{\text{total}}$: Total number of trainable parameters in the model.
    \item $P_{\text{block}_i}$: Number of parameters in block $i$.
    \item $B$: Bytes per parameter (e.g., $4$ for FP32, $2$ for FP16).
    \item $k$: Percentage of top blocks selected for training.
    \item $S_k$: Set of blocks in the top $k$.
\end{itemize}

\noindent For full fine-tuning, the optimizer states for all parameters reside on the GPU:
\[
\text{Mem}_{\text{Full}} = 2 \times P_{\text{total}} \times B
\]

\noindent For our selective method, only the optimizer states for the selected blocks reside on the GPU. The number of parameters in these blocks is:
\[
P_{\text{selected}} = \sum_{i \in S_k} P_{\text{block}_i}
\]
and the GPU memory usage becomes:
\[
\text{Mem}_{\text{Selective}} = 2 \times P_{\text{selected}} \times B
\]

\noindent The absolute memory saved is:
\[
\text{Mem}_{\text{Saved}} = \text{Mem}_{\text{Full}} - \text{Mem}_{\text{Selective}} = 2 \times (P_{\text{total}} - P_{\text{selected}}) \times B
\]

\noindent The percentage reduction in optimizer state memory is:
\[
\% \text{Reduction} = \left( 1 - \frac{P_{\text{selected}}}{P_{\text{total}}} \right) \times 100
\]
This formula provides a deterministic way to calculate GPU memory savings achieved by the selective optimizer state residency approach.

\section{Experiments Setting}
To evaluate the effectiveness of the proposed approach, a series of experiments were designed. The following subsections detail the objectives and the setup of these experiments.
\subsection{Experimental Goals}

The primary objectives of our experiments are to evaluate the effectiveness and efficiency of \textbf{AdaGradSelect} compared to baseline fine-tuning approaches, with a focus on model performance, convergence behavior, and training efficiency. Specifically, we address the following research questions:

\subsubsection*{\textbf{Q1) Performance and Generalization:}}  
How does \textbf{AdaGradSelect} compare to existing methods such as \textit{LoRA} and full fine-tuning in terms of task performance, and does it maintain consistent generalization across different LLM architectures and model size increase ?

\subsubsection*{\textbf{Q2) Loss Convergence:}}  
How does \textbf{AdaGradSelect} influence the convergence behavior of training compared to LoRA and full fine-tuning? 

\subsubsection*{\textbf{Q3) Training Efficiency:}}  
To what extent does \textbf{AdaGradSelect} improve training efficiency, particularly in terms of GPU memory reduction and overall compute savings ?

\begin{figure}[H]
    \centering
    \includegraphics[width=\columnwidth]{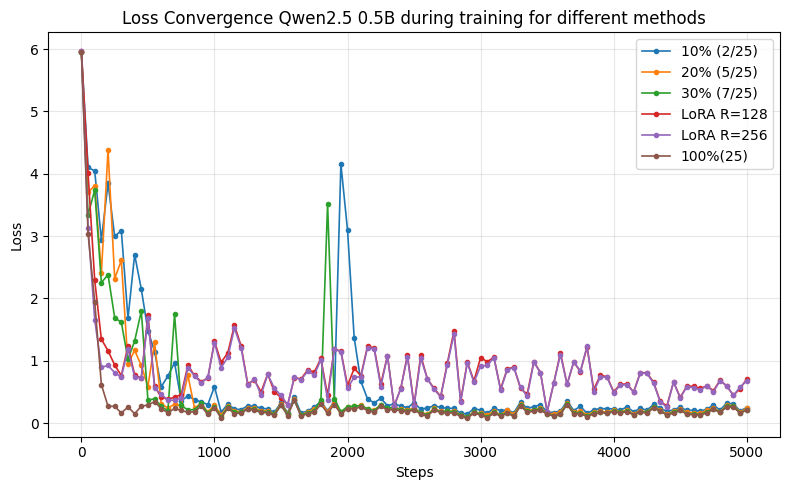}
    \caption{Loss convergence of Qwen2.5 0.5B on MetaMath40K for AdaGradSelect(10-30\%) and other methods }
    \label{fig:loss_convergence}
\end{figure}

\subsection{Experimental Setup}

We evaluate \textbf{AdaGradSelect} on mathematical reasoning tasks using three models: \textit{Qwen2.5-0.5B}, \textit{LLaMA-3.2-1B}, and \textit{Phi-4 Mini-3.8B}. The models are fine-tuned on the \textit{MetaMathQA-40K} dataset \cite{yu2024metamathbootstrapmathematicalquestions} and evaluated on \textit{GSM8K} \cite{cobbe2021trainingverifierssolvemath} and \textit{MATH} \cite{hendrycks2021measuringmathematicalproblemsolving}. These benchmarks assess the mathematical reasoning capabilities added through fine-tuning, since the base models lack such ability due to their small size.

For a fair comparison, we follow the setup from \cite{lingam2024svftparameterefficientfinetuningsingular}, evaluating AdaGradSelect, LoRA , and Full Fine-tuning. LoRA adapts Q, K, V, U, D, O, and G projections with ranks $r=128$ and $r=256$, standard in prior works. For AdaGradSelect, we use 10\%–30\% block update variants, which balance performance and efficiency. All models are trained with \textbf{BF16 precision}.

All evaluations are conducted in a zero-shot setting without system prompts. We use greedy decoding (\textit{temperature = 0}) to produce deterministic outputs; hence, no variance across runs is reported.

\noindent \textbf{Hardware:} All experiments are conducted on a Linux server equipped with a single NVIDIA RTX A6000 GPU (48 GB VRAM), connected via PCIe Gen 4.0 $\times$16.

\begin{table*}[h!]
\centering
\caption{Performance of AdaGradSelect, LoRA, and Full Fine-Tuning accuracies on GSM8K and MATH across different models. Best performances are highlighted in bold, while the second-best
performances are underlined.}
\label{tab:merged-performance}
\renewcommand{\arraystretch}{1.2}
\setlength{\tabcolsep}{6pt}

\begin{tabular}{lcccccc}
\toprule
\textbf{Method} & \multicolumn{2}{c}{\textbf{Qwen2.5-0.5B}} & \multicolumn{2}{c}{\textbf{LLaMA3.2-1B}} & \multicolumn{2}{c}{\textbf{Phi4-mini-3.8B}} \\
\cmidrule(lr){2-3} \cmidrule(lr){4-5} \cmidrule(lr){6-7}
 & \textbf{GSM8K} & \textbf{MATH} & \textbf{GSM8K} & \textbf{MATH} & \textbf{GSM8K} & \textbf{MATH} \\
\midrule
\textbf{AdaGradSelect (10\%)} & 50.72 & 27.72 & 50.11 & 28.92  & 84.53 & 46.03 \\
\textbf{AdaGradSelect (20\%)} & 51.18 & 26.52 & 51.18 & 27.32 & \textbf{85.90} & 47.29  \\
\textbf{AdaGradSelect (30\%)} & \textbf{52.39} & \underline{27.86} & \textbf{54.97} & \underline{31.60}  & 85.44 & 46.69  \\
\midrule
\textbf{LoRA (r=128)} & 50.26 & \textbf{27.92} & \underline{54.81}  & 30.86 & \underline{83.55} & \underline{46.83} \\
\textbf{LoRA (r=256)} & 50.37 & 27.12 & 52.69 & 29.46  &  83.24 &  \textbf{47.76} \\
\midrule
\textbf{Full Fine-Tuning} & \underline{51.47} & 26.19 & 54.35  & \textbf{33.07}  & 85.37 & \textbf{47.76}   \\
\bottomrule
\end{tabular}
\end{table*}



\section{Results and Discussion}
This section presents the experimental results, analyzes the performance of our method and discusses its efficiency and convergence behaviour.
\subsection{Performance and Generalization Across Models}

Table~\ref{tab:merged-performance} reports the performance of AdaGradSelect, LoRA, and full fine-tuning across three model families: Qwen2.5-0.5B, LLaMA3.2-1B, and Phi4-mini-3.8B, evaluated on GSM8K and MATH benchmarks.  

\textbf{Performance Comparison:}  
It can be observed that AdaGradSelect achieves performance highly comparable to full fine-tuning and consistently surpasses LoRA, even with larger rank settings such as $r=256$. For example, on Qwen2.5-0.5B (GSM8K), AdaGradSelect (30\%) reaches 52.39, outperforming LoRA (r=256) at 50.37, while remaining competitive with full fine-tuning at 51.47. Similarly, on LLaMA3.2-1B, AdaGradSelect (30\%) achieves 54.97 on GSM8K, exceeding both LoRA variants and even slightly surpassing full fine-tuning (54.35). These results indicate that AdaGradSelect provides a strong trade-off, delivering near full fine-tuning performance with substantially reduced training costs.  

\textbf{Generalization Across Models:}  
A notable strength of AdaGradSelect is its robustness across model scales and architectures. In the case of \textbf{LLaMA3.2-1B}, which contains only 18 transformer blocks compared to Qwen2.5-0.5B with 25 blocks, the 10\% setting corresponds to updating only a single block per iteration. Despite this extreme sparsity, performance degradation remains modest (50.11 GSM8K, 28.92 MATH), demonstrating that even minimal block updates can drive meaningful adaptation. This property underlines the method’s efficiency, particularly in smaller models where parameter updates are more constrained.  

Moreover, results across the larger \textbf{Phi4-mini-3.8B} confirm that scaling the model does not diminish AdaGradSelect’s competitiveness. At 30\%, AdaGradSelect achieves 85.44 (GSM8K) and 46.69 (MATH), closely matching full fine-tuning (85.37 and 47.76) while outperforming LoRA under both rank configurations.  

\textbf{Guideline for Block Selection:}  
Our experiments show that increasing the percentage of updated blocks generally improves performance, consistent with expectations. An important guideline emerges for practitioners: at least one transformer block should be updated in each iteration. Formally, for a model with $B$ blocks, the minimum selection percentage should satisfy:
\[
\text{min\%} \geq \frac{1}{B} \times 100,
\]
ensuring that every iteration updates at least one block. This simple rule provides a practical lower bound for setting the AdaGradSelect hyperparameter.  

In summary, AdaGradSelect demonstrates strong generalization across diverse models and tasks, offering near full fine-tuning accuracy while surpassing LoRA in both performance and stability. This establishes AdaGradSelect as a promising alternative for efficient fine-tuning in large language models.




\subsection{Loss Convergence}

Figure~\ref{fig:loss_convergence} shows the loss convergence behavior of Qwen2.5 0.5B across different fine-tuning strategies, including partial parameter selection with AdaGrad (10\%, 20\%, 30\%), LoRA with varying rank (128, 256), and full fine-tuning (100\%).  

All methods exhibit a sharp initial decrease in loss, indicating effective adaptation during early training. For AdaGradSelect, convergence is initially slower than full fine-tuning, but the gap narrows as training progresses. With higher selection ratios (20\% and 30\%), AdaGradSelect eventually achieves convergence behavior comparable to full fine-tuning, though with slightly higher variance.  

In contrast, LoRA with both rank 128 and 256 demonstrates consistently slower convergence compared to AdaGradSelect and full fine-tuning. The two LoRA curves largely overlap, reflecting similar behavior, and both exhibit higher variance and reduced stability. This is a consequence of low-rank adaptation: while LoRA achieves efficiency gains by reducing trainable parameters, its optimization is constrained to a low-rank subspace, which limits representational power and prevents it from fully matching the performance of full fine-tuning.  

Overall, full fine-tuning provides the most stable and lowest loss trajectory, while AdaGradSelect offers a promising trade-off between efficiency and convergence. LoRA, although parameter-efficient, converges more slowly and remains constrained by its low-rank parameterization.


\subsection{Training Efficiency Comparison}

Figure~\ref{fig:gpu_vs_time} compares the usage of GPU and training time for different methods. AdaGradSelect demonstrates a significant reduction in optimizer state memory compared to full fine-tuning, as it updates only a fraction of parameters (10\%, 20\%, 30\%). This selective update strategy yields substantial memory savings while still maintaining competitive convergence, highlighting its suitability for resource-constrained environments.

LoRA also provides memory efficiency by restricting adaptation to low-rank matrices. One could argue that using smaller ranks, such as $r=64$ or $r=32$, would further reduce GPU memory consumption. However, prior work and empirical evidence suggest that such low ranks lead to a marked degradation in performance, as the adaptation subspace becomes too limited to capture task-specific information effectively. In practice, $r=128$ and $r=256$ are widely adopted values for LoRA, striking a balance between memory efficiency and model adaptability.

Compared to these, AdaGradSelect offers a more flexible and advantageous trade-off. By adjusting the percentage of parameters selected, it achieves greater GPU memory reduction than LoRA at commonly used ranks (128 and 256), while avoiding the significant performance penalties associated with very low ranks. This makes AdaGradSelect particularly attractive for efficient fine-tuning of small-scale models under strict memory budgets.

\section{Limitations}
While the dynamic optimizer state management strategy in AdaGradSelect significantly reduces GPU memory overhead, it introduces a dependency on the PCIe bandwidth of the underlying hardware. Efficient prefetching and eviction of optimizer states rely on high transfer rates between CPU and GPU memory. On systems with slower interconnects, this could become a performance bottleneck, potentially negating some of the gains from reduced GPU usage. In our experiments with small-to-medium language models (SLMs), we did not encounter noticeable slowdowns, but scaling to larger models or deploying on hardware with limited PCIe throughput may present challenges. Future work could explore overlap techniques or NVLink-based setups to further mitigate transfer latency.

\section{Conclusion}
We introduced AdaGradSelect, an adaptive gradient-guided block selection strategy for efficient fine-tuning of language models. Across three small model families and two reasoning benchmarks, AdaGradSelect consistently matches or surpasses LoRA while approaching the accuracy of full fine-tuning at a fraction of the cost. Its robustness holds across both small and large models, with even small number of block updates (e.g., 10\%) yielding meaningful adaptation.

Experiments show that AdaGradSelect converges faster and more stably than LoRA, while substantially reducing memory and training time compared to full fine-tuning.

In general, AdaGradSelect offers a compelling trade-off between performance and efficiency, positioning it as a strong alternative to full fine-tuning and low-rank adaptation methods for resource-constrained SLM / LLM training.

\bibliographystyle{ACM-Reference-Format}
\bibliography{ref}


\end{document}